\documentclass[sigconf,nonacm]{acmart}
\makeatletter
\setkeys{acmart.cls}{pbalance=true}
\makeatother
\AtBeginDocument{  }

\setcopyright{none}
\usepackage[vlined,linesnumbered, ruled]{algorithm2e}
\usepackage{xcolor}

\usepackage{ifthen}

\usepackage{amsmath}

\usepackage{amssymb}
\usepackage{tikz-cd}
\usepackage{cleveref}
\SetKwRepeat{Do}{do}{while}\SetKwIF{WithProb}{ElseWithProb}{Else}{with probability}{do}{else with probability}{else}{end}

\newcommand{\realnum}{\mathbb{R}}
\newcommand{\natnum}{\mathbb{N}}

\newcommand{\SphereOneMax}{\textsc{SphereOneMax}\xspace}

\newcommand{\OneMax}{\textsc{OneMax}\xspace}
\newcommand{\Sphere}{\textsc{Sphere}\xspace}

\newcommand{\GenSphereLinear}{\textsc{GeneralizedProximityOneMax}\xspace}
\newcommand{\MCProximity}{\textsc{MCProximity}\xspace}
\newcommand{\WPMCProximity}{\textsc{ProximityOneMax}\xspace}
\newcommand{\LexicoSphereInt}{\textsc{LexicoSphereInt}\xspace}

\newcommand{\oneoneEA}{(1+1)-EA\xspace}
\newcommand{\oneoneES}{(1+1)-ES\xspace}
\newcommand{\oneoneEAESBL}{(1+1)-EA-ES-Bilevel\xspace}
\newcommand{\oneoneEAES}{(1+1)-EA-ES\xspace}
\newcommand{\oneoneEAESHB}{(1+1)-EA-ES-Integrated\xspace}

\newcommand{\Ex}[1]{\mathrm{E}\left[#1\right]}
\newcommand{\Prob}[1]{\mathrm{Pr}\left[#1\right]}
\newcommand{\BigO}[1]{\mathcal{O}\left(#1\right)}

\newcommand{\Brackets}[1]{\mathopen{}\left(#1\right)\mathclose{}}

\newcommand{\CurlyBrackets}[1]{\left \{ #1 \right \}}
\newcommand{\DoubleVert}[1]{\left \lvert\left \lvert #1 \right \rvert\right \rvert}

\newcommand{\maxWBrackets}[1]{\max\CurlyBrackets{#1}}
\newcommand{\minWBrackets}[1]{\min\CurlyBrackets{#1}}
\newcommand{\logWBrackets}[1]{\log\Brackets{#1}}
\newcommand{\ThetaWBrackets}[1]{\Theta\Brackets{#1}}

\renewcommand{\epsilon}{\varepsilon}

\newcommand{\Normal}{\mathcal{N}}

\newcommand{\firstInts}[1]{[#1]}
\newcommand{\NormTwo}[1]{\DoubleVert{#1}^2}
\newcommand{\NormEucl}[1]{\DoubleVert{#1}}
\newcommand{\NormUnderscoreOne}[1]{\DoubleVert{#1}_1}
\newcommand{\NormUnderscoreZero}[1]{\DoubleVert{#1}_0}

\AtBeginDocument{      }

\begin{document}

\title{
Analysis of the (1+1)-EA-ES-Bilevel on Mixed Discrete-Continuous Optimization Problems with Interactions
}

\author{Timo Kötzing}
\email{timo.koetzing@hpi.de}
\orcid{0000-0002-1028-5228}
\affiliation{  \institution{Hasso Plattner Institute\\ University of Potsdam}
  \city{Potsdam}
  \country{Germany}
}

\author{Jurek Sander}
\email{jurek.sander@hpi.de}
\orcid{0009-0004-6707-6592}
\affiliation{  \institution{Hasso Plattner Institute\\ University of Potsdam}
  \city{Potsdam}
  \country{Germany}
}

\renewcommand{\shortauthors}{Kötzing and Sander}

\begin{abstract}
Mixed discrete-continuous optimization problems can pose challenges -- especially
for previous integrated approaches -- when interactions cause beneficial
discrete changes to be rejected before sufficient continuous adjustment.

We introduce \WPMCProximity as a simple benchmark with pairwise
discrete-continuous interactions.
On this function, we prove an exponential lower bound on the
expected time for the \oneoneEAESHB to reach the target accuracy
when initialized near a continuous optimum for a suboptimal
discrete configuration with a sufficiently small step size.
Experiments show frequent entries into such states from random initialization.

We introduce the \oneoneEAESBL, which uses an upper-level \oneoneEA
to optimize discrete variables and a lower-level \oneoneES to optimize
continuous variables for each fixed discrete offspring before selection.
For several interaction benchmarks and sufficiently small target
values $\epsilon>0$ that require an optimal discrete configuration,
using fixed continuous initialization and a
suitable fixed budget per lower-level call, we prove that the expected
number of fitness evaluations to reach objective value at most
$\epsilon$ is $\ThetaWBrackets{n_d \cdot \log n_d \cdot n_c \cdot \logWBrackets{R/\sqrt{\epsilon}} }$, where $n_d$ is the number of discrete variables, $n_c$ is the number of continuous variables, and $R$ bounds the distances from the fixed continuous starting point to all relevant continuous optima for fixed discrete configurations.
\end{abstract}



\maketitle
\hypersetup{pdfauthor={Timo Kötzing and Jurek Sander}}

\section{Introduction}
Many real-world optimization problems combine combinatorial decisions with continuous operating parameters, thereby involving both discrete and continuous variables \cite{mixed1,mixed2,mixed4,talbi2024mixedsurvey}.
These variables are often interdependent.
In particular, changing the discrete configuration can determine which continuous variables are relevant and which continuous values are optimal \cite{icatcma_akimoto}.
Optimizing such \emph{mixed discrete-continuous} problems with \emph{interactions} therefore requires coordinating the search over both variable types \cite{li2013mies}.

For example, in a network connection problem \cite{network3}, discrete decisions determine whether relay stations are built, while continuous variables determine signal strengths.
Building a relay station changes the signal strengths required to maintain connectivity; its benefit, therefore, depends on adapting the continuous variables to the new discrete configuration.
Such a change can appear detrimental until the continuous parameters have been sufficiently adjusted~\cite{icatcma_akimoto,igbd_akimoto}.

To theoretically study the algorithmic difficulties caused by these interactions, we consider benchmark problems that isolate these mechanisms \cite{gambitThesis,icatcma_akimoto,igbd_akimoto}.
We focus on the mixed search space $\{0,1\}^{n_d} \times \realnum^{n_c}$ with $n_d$ discrete and $n_c$ continuous variables.
While our benchmarks also cover settings with different numbers of discrete and continuous variables, the difficulty of coordinating the two variable types already arises when each discrete variable interacts with a single continuous variable.
To illustrate this, we consider the case $n_d = n_c = N$ and introduce the \WPMCProximity benchmark, which is based on the work of \cite{catcma_hamano,gambitThesis}.
For a weight $w>0$, its objective function is defined as
\begin{align*}
 (x,y) \in \{0,1\}^{N} \times \realnum^{N}  &\mapsto \sum_{i=1}^{N}\Brackets{\Brackets{\Brackets{1-x_i} - y_i}^2 + w \Brackets{1-x_i}}.
\end{align*}
For each $i \in \firstInts{N}$, the quadratic term measures the
deviation of $y_i$ from $1-x_i$, its optimal value when $x_i$
is fixed, which we call its \emph{conditional optimum}.
Thus, changing $x_i$ shifts the optimal value of $y_i$.
The term $w\Brackets{1-x_i}$ penalizes $x_i=0$, so the unique global optimum is $\Brackets{x^*,y^*} = \Brackets{1^N,0^N}$.

This benchmark captures the need for continuous adjustment illustrated by the network connectivity example.
To see this, consider a single pair at $\Brackets{x_i,y_i}=(0,1)$, where the continuous variable is optimal for the current bit value.
For $0<w<1$, flipping $x_i$ to~$1$ while keeping~${y_i=1}$ increases the pair's contribution from $w$ to~$1$.
Adjusting $y_i$ to its new conditional optimum~$0$ then reduces this contribution to~$0$.
Thus, the benefit of the discrete change becomes apparent only after sufficient continuous adjustment.
An integrated algorithm that immediately selects based on the joint objective may therefore reject the discrete change if the accompanying continuous mutation does not sufficiently reduce this error.

Following \cite{anne2027mixeddiscretecontinuous}, we call a mixed discrete-continuous problem \emph{separable} if it is \emph{ordinally separable}, meaning that the ranking of solutions within either variable subspace is independent of the fixed values in the other subspace.
In particular, the set of minimizers within either subspace is independent of these fixed values.
The \WPMCProximity benchmark is not separable in this sense, since the optimal continuous vector is $y=1-x$ and thus depends on the discrete configuration.
However, the objective is a sum of independent terms, each involving only one discrete and one continuous variable.
It therefore isolates a minimal form of interaction between the two variable types.

Several evolution-strategy-based approaches \cite{schwefel1984evolution} have been proposed and empirically evaluated for mixed discrete-continuous optimization.
Building on the CMA-ES \cite{hansen2001cmaes}, Hamano et al.\ \cite{catcma_hamano} introduced the CatCMA, which samples continuous variables from a multivariate Gaussian distribution and discrete variables from a categorical distribution.
CatCMA with Margin extends this framework to jointly handle continuous, integer, and categorical variables~\cite{hamano2025catcmawm}.
Akimoto et al.~\cite{icatcma_akimoto} showed that CatCMA struggles with interactions between discrete and continuous variables.
Their extension, ICatCMA, combines warm-starting with a parameterized affine map from the categorical to the continuous space to represent these dependencies within the joint distribution.
In contrast to these \emph{integrated} approaches, a \emph{bilevel} approach organizes the search into two nested levels.
In the setting considered here, the upper level searches over discrete configurations, while the lower level optimizes the continuous variables for each proposed configuration, keeping its discrete variables fixed.
The upper level then evaluates the proposed configuration using the resulting joint objective value.
Ong et al.\ \cite{igbd_akimoto} identified interaction settings that remain challenging for ICatCMA in their experiments and proposed the IGBD algorithm, which uses such a bilevel approach.
This allows continuous adjustments to reveal the benefit of a discrete change before upper-level selection.
This nested structure introduces a trade-off between the accuracy of conditional continuous optimization and the number of fitness evaluations spent per discrete candidate.
Independent of mixed discrete-continuous optimization, applying evolutionary algorithms to the bilevel approach has been studied in various contexts, e.g., in \cite{bilevel_cont_akimoto,akimoto2027evolutionarybilevel,chen2025bilevel1,chen2024bilevel2,he2019bilevel3,huang2023bilevel4}.
Beyond evolution strategies, GAMBIT explicitly learns and exploits dependencies between discrete and continuous variables within a model-based evolutionary algorithm~\cite{sadowski2018gambit}.
Another hybrid approach combines CMA-ES with population-based incremental learning for the binary variables and has been evaluated empirically on separable and non-separable mixed-binary benchmarks~\cite{kotzing2025mixedbinary}.

In contrast, less is known about the theoretical performance of algorithms for mixed discrete-continuous optimization problems.
Earlier work on mixed-integer evolution strategies established global convergence under suitable assumptions~\cite{li2013mies}.
Hamano et al.~\cite{hamano2026convergence} analyze the (1+1)-LB-ES and (1+1)-LUB-ES on \LexicoSphereInt, a separable lexicographically ordered mixed-integer sphere benchmark, but consider only the phase in which the integer variables are already optimal.
Auger et al.\ \cite{anne2027mixeddiscretecontinuous,anne2027theoretical} proposed the \oneoneEAES with reduced success rule, which is a hybrid integrated algorithm that treats the discrete and continuous variables with their respective optimization strategies.
A companion run time analysis \cite{anne2027theoretical} covers the full optimization process and establishes efficient optimization for the separable \SphereOneMax benchmark.

In this paper, we provide a theoretical run time analysis of integrated and bilevel search on mixed discrete-continuous benchmarks with interactions between the two variable types.
We establish conditions under which the integrated approach requires exponential expected time and prove that a simple bilevel approach optimizes the same benchmark in polynomial expected time.

Our contributions and the organization of the paper are as follows.
In \Cref{sect:prelims}, we introduce the notation and theoretical tools used in the analysis, and in \Cref{sect:problem-desc}, we present a list of benchmark problems and their interaction mechanisms.
This includes the separable function \SphereOneMax, the three types of interactions (Type-I - Type-III) between discrete and continuous variables as introduced in \cite{icatcma_akimoto}, and the \WPMCProximity function.
Furthermore, we define the function class \GenSphereLinear, which provides a common framework for several of these benchmarks.

In \Cref{sect:analysis-integrated}, we identify a limitation of the integrated \oneoneEAES with reduced success rule on \WPMCProximity for $0<w<1$.
We prove an exponential lower bound on the expected time to reach the target accuracy when the algorithm starts
sufficiently close to a suboptimal conditional continuous optimum with a sufficiently small step size.
Experiments show frequent entries into these trap states starting from random initializations, even for small numbers of dimensions.

To overcome the difficulties of the integrated approach, we introduce a bilevel approach in \Cref{sect:algorithm}, similar to \cite{igbd_akimoto}, that allows the algorithm to optimize the continuous variables for a proposed discrete configuration before deciding whether to accept the combined offspring.
We focus on the $(1+1)$ setting to obtain a simple but standard framework for analyzing the interaction between discrete mutation, continuous adaptation, and selection.
The resulting \oneoneEAESBL combines the \oneoneEA at the upper level with the \oneoneES at the lower level.
In the configuration analyzed here, every lower-level call starts at the same continuous point, with a fixed initial step size, and receives a fixed budget of fitness evaluations.
The remaining continuous error can affect the ranking of discrete candidates, making upper-level selection nontrivial to analyze.
To handle this complexity in the following analysis, we utilize the fixed-budget success guarantee derived in \Cref{sect:fixed-budget-guarantees}, which holds uniformly over the relevant dimensions and admissible initial points.

In \Cref{sect:analysis-bilevel}, we establish upper and lower bounds of the same asymptotic order for \GenSphereLinear and derive results for \SphereOneMax, Type-II interactions, and \WPMCProximity.
We additionally obtain corresponding bounds for restricted Type-I and Type-III benchmarks.
The analyzed benchmarks isolate coordination effects through translated spherical subproblems, possibly restricted to active
coordinates.
Exact continuous minimization leaves a scaled \OneMax objective; the analysis quantifies the effect of finite randomized lower-level optimization on the joint search.
To obtain the bounds, the lower-level budget requires an available upper bound~$R$ on the distance from the fixed continuous starting point to every relevant conditional optimum.
Any valid upper bound suffices, and its influence on the budget and run time bound is logarithmic.
With this budget rule based on the distance bound $R$ and $\epsilon$ as target objective accuracy, we show that the expected number of fitness evaluations on all analyzed benchmark variants is in $\ThetaWBrackets{n_d \cdot \log n_d \cdot n_c \cdot \logWBrackets{R/\sqrt{\epsilon}} }$.

Our analysis uses drift theory to bound the expected optimization time \cite{kötzing2024theorystochasticdrift}.
The central difficulty is that finite randomized lower-level runs provide stochastic evaluations of the conditionally optimized objective, connecting our analysis to run time analyses of evolutionary algorithms with noisy fitness evaluations \cite{giessen2016robustness}.
In our setting, the evaluation error is the nonnegative residual from continuous optimization, whose distribution may depend on the discrete configuration and the lower-level budget.
These errors can change the ranking of discrete candidates and thereby affect upper-level selection.
We retain previously obtained fitness values without reevaluating the current solution, a choice that has also been studied in noisy evolutionary optimization \cite{DoerrNoise2025}.
We control the resulting errors through a uniform success guarantee for the lower-level optimizer and a potential function that accounts for the continuous residual error in the upper-level process.
Elitist selection ensures that the potential cannot increase, even when a lower-level run fails to reach the required accuracy.
Combining this monotonicity with the probability of sufficiently accurate lower-level optimization yields a multiplicative drift bound and hence the run time upper bound.
A comparison with the discrete mutation process analyzed by Witt \cite{WITT_2013}, together with the fixed evaluation budget of each lower-level call, yields the lower bound.
\section{Preliminaries}
\label{sect:prelims}

We write $\realnum_+=[0,\infty)$ and $\firstInts{n}=\{1,\ldots,n\}$ for $n\in\natnum$.\newline
For~${x\in\{0,1\}^{n_d}}$, we use $\NormUnderscoreZero{x}$ and $\NormUnderscoreOne{x}$ to count zero and one bits, respectively.
All logarithms are natural and $\natnum$ includes~$0$.

For the run time lower bound of the \oneoneEAESBL, we use the following result of Witt~\cite{WITT_2013}, which gives an
asymptotically tight lower bound for the \oneoneEA on linear functions, including \OneMax.

\begin{theorem}
\label{thm:witt-runtime-linear}
    Let $f$ be a linear fitness function and consider the \oneoneEA optimizing $f$.
    Let $T$ denote the number of iterations until an optimum is reached, and $n_d$ be the number of variables.
    Assume that all coefficients of $f$ are nonzero and that the initial bit string is sampled uniformly at random.
    On any linear function, the optimization time of the \oneoneEA with mutation probability $1/n_d$ is bounded from below by
    \begin{align*}
        \Ex{T} \geq (1-o(1))\cdot e \cdot n_d \cdot \ln\Brackets{n_d}.
    \end{align*}
\end{theorem}

To obtain the upper bound of the run time of the \oneoneEAESBL we use drift analysis, originally introduced by He and Yao~\cite{HE_yao_drift_2001, he_yao_drift_study_2004}.
See \cite{kötzing2024theorystochasticdrift} for a general overview of the technique.
We require the following multiplicative drift theorem as given in~\cite{doerr2010multiplicative}.

\begin{theorem}
Let $\Brackets{X_t}_{t\ge 0}$ be an integrable process with values in
$\{0\}\cup[1,\infty)$, and let
$T=\inf\{t\ge 0:X_t=0\}$.
Assume that there is a constant $\delta>0$ such that, for all $t<T$,
\(
\Ex{X_t-X_{t+1}\mid X_0,\ldots,X_t}\ge \delta X_t.
\)
Then
\[
\Ex{T}
\le \frac{1+\ln\Ex{X_0}}{\delta}.
\]
    \label{thm:prelim:mult}
\end{theorem}
\section{Problem Description}
\label{sect:problem-desc}

In this section, we collect and generalize definitions of benchmark functions to evaluate mixed discrete-continuous algorithms; this list includes the separable function \SphereOneMax, the three types of interactions (Type-I - Type-III) between discrete and continuous variables as defined in \cite{icatcma_akimoto}, and the \WPMCProximity function which we introduce as a weighted variant of the MixedCategoryProximity (\MCProximity) function in~\cite{catcma_hamano} and the function $F_6$ in~\cite[Chapter~4]{gambitThesis}.
Additionally, we define the function class \GenSphereLinear to demonstrate the similarities among most of the benchmark functions.

We minimize all objectives and seek a solution whose objective value is within $\epsilon>0$ of the global minimum.
Gaussian mutation hits a continuous optimum with probability zero.

Let $n_d,n_c\in \natnum$ be the number of discrete and continuous variables, $x \in \{0,1\}^{n_d}$ the discrete bit string that gets optimized and~$y\in\realnum^{n_c}$ the continuous variables.
For each $x\in\{0,1\}^{n_d}$, let $y^*(x)$ denote a chosen conditional continuous minimizer.
Such a minimizer need not be unique.

\paragraph{\WPMCProximity}
As already defined in the introduction, for $n_d=n_c=N\in \natnum$, the \WPMCProximity function creates pairwise interactions between the discrete and continuous variables.
Let $w > 0$ be a fixed weight.
Then, \WPMCProximity is given as the sum of all pairwise connections by
\begin{align*}
 (x,y) \in \{0,1\}^{N} \times \realnum^{N}  &\mapsto \sum_{i=1}^{N}\Brackets{\Brackets{\Brackets{1-x_i} - y_i}^2 + w \Brackets{1-x_i}}.
\end{align*}
The difficulty of optimizing the function is controlled by the dimensionless parameter $w$.
For $w = 0.5$, it is equivalent to the benchmark function $F_6$, defined in~\cite[Chapter~4]{gambitThesis}.

\begin{figure}[htp]
    \centering
    \includegraphics[width=\columnwidth]{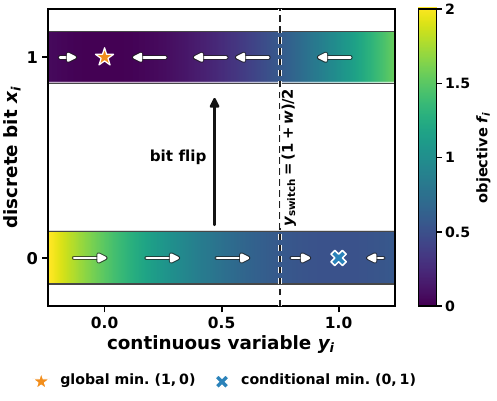}
    \caption{Pairwise landscape for~${w=0.5}$. The two bands correspond to $x_i\in\{0,1\}$, the color represents the objective value, and the arrows indicate the negative gradient with respect to~$y_i$. A flip from $x_i=0$ to $x_i=1$ shifts the continuous optimum from $y_i=1$ to $y_i=0$. The dashed line marks the equal-value threshold $y_{\mathrm{switch}}=(1+w)/2$.}
    \label{fig:wpmcproximity-landscape}
    \Description{Two horizontal heat-map bands are shown over the continuous variable $y_i$. The lower band corresponds to $x_i=0$ and has its minimum at $y_i=1$, whereas the upper band corresponds to $x_i=1$ and has its minimum at $y_i=0$. Horizontal arrows point toward the respective continuous minima. A vertical arrow illustrates a bit flip at fixed $y_i$, which changes the direction of continuous optimization. A dashed vertical line at $y_i=0.75$ marks the point at which both bit values have equal objective value. The conditional minimum $\Brackets{x_i,y_i}=(0,1)$ and the global minimum $\Brackets{x_i,y_i}=(1,0)$ are highlighted by separate markers.}
\end{figure}

\Cref{fig:wpmcproximity-landscape} shows the contribution of one pair $\Brackets{x_i,y_i}$. For $x_i=0$, continuous optimization drives $y_i$ toward $1$,
whereas flipping $x_i$ to $1$ shifts the conditional optimum to $0$. For $0<w<1$, flipping the bit while keeping~${y_i=1}$ increases the objective from $w$ to $1$, creating a transition barrier of $1-w>0$. Reaching the global optimum $(1,0)$ therefore requires coordinated progress in both variable domains.

\paragraph{\GenSphereLinear}

We define this fitness function class as follows.
Let $y^*: \{0,1\}^{n_d} \rightarrow \realnum^{n_c}$ be an arbitrary mapping, and let $w >0$ be a fixed constant.
Then, a function of the class \GenSphereLinear is given as
\begin{align*}
 (x,y) \in \{0,1\}^{n_d} \times \realnum^{n_c}  \mapsto \sum_{i=1}^{n_d} w\Brackets{1-x_i} + \NormTwo{y - y^*(x)}.
\end{align*}
For each fixed discrete bit string $x$, the problem is a spherical function in $y$ with the unique optimum $y^*(x)$.
The global optimum is given by $\Brackets{x^*, y^*}=\Brackets{1^{n_d},y^*\Brackets{1^{n_d}}}$.

This function class generalizes multiple test function classes. \Cref{tab:functions:gen-sphere-lin} gives the parameters for \SphereOneMax, Type-II, and \WPMCProximity as functions of this class.
As the standard benchmark fitness function, where the discrete and continuous variables can be solved separately, we use \SphereOneMax.
The Type-II interaction, as introduced in \cite{icatcma_akimoto}, characterizes cases in which the discrete bit string determines the optimal solution of the continuous variables using a mapping $\phi^*: \{0,1\}^{n_d} \rightarrow \realnum^{n_c}$.
With $V^* \in \realnum^{n_c\times n_d}$, $b^* \in \realnum^{n_c}$ and $a\geq 0$, where the parameter~$a$ controls the strength of the discrete-continuous interaction, we define for $x \in \{0,1\}^{n_d}$ the mapping $\phi^*$ as $\phi^*(x) = aV^*x+b^*$.
For $a\rightarrow 0$, the problem approaches separability.

\begin{table}[htp]
    \centering
    \captionsetup{position=bottom}
    \begin{tabular}{@{}lll@{}}
        \toprule
        Benchmark & $w$ & $y^*(x)$ \\
        \midrule
        \SphereOneMax \cite{gambitThesis} & $1$ & $0^{n_c}$ \\
        Type-II \cite{icatcma_akimoto} & $1$ & $\phi^*(x)=aV^*x+b^*$ \\
        \WPMCProximity & $w>0$ & $1-x$ \\
        \bottomrule
    \end{tabular}
    \caption{Specializations of \GenSphereLinear. For \WPMCProximity, $n_d=n_c=N$.}
    \label{tab:functions:gen-sphere-lin}
\end{table}

\paragraph{Type-I and Type-III Interactions}
The following two functions are originally introduced in \cite{icatcma_akimoto} and not part of the \GenSphereLinear class.
The Type-I interaction characterizes scenarios in which the discrete bit string $x$ determines which components of the continuous variables $y$ are active.
The Type-III interaction combines this masking with the dependence of the continuous optimum on the discrete bit string.
Assume $n_d=n_c=N$, let $b^*\in\realnum^{n_c}$ be a fixed vector that does not contain zeros for Type-I, and let~$\phi^*$ be the affine mapping defined above for Type-III.
We denote the Hadamard product by~$\odot$.
Then, we define the functions as
\begin{align*}
    f_I(x,y) &= \NormTwo{x\odot y-b^*},\\
    f_{III}(x,y) &= \NormTwo{x\odot y-\phi^*(x)}.
\end{align*}
For fixed $x$, the vectors $b^*$ and $\phi^*(x)$, respectively, are conditional minimizers; coordinates with $x_i=0$ are arbitrary.
A global optimum is given by $\Brackets{1^N,b^*}$ for Type-I and $\Brackets{1^N,\phi^*\Brackets{1^N}}$ for Type-III; for Type-I, it is unique since $b^*$ contains no zeros.
These objectives do not belong to \GenSphereLinear, because the discrete variables determine which continuous coordinates are active.
\section{Limitations of the Integrated Approach}
\label{sect:analysis-integrated}

Before introducing the \oneoneEAESBL and analyzing its performance, we first show that an integrated approach that operates on just one level, namely the \oneoneEAES with reduced success rule as defined and analyzed in \cite{anne2027mixeddiscretecontinuous,anne2027theoretical}, can require exponential expected optimization time on problems with interactions.
This occurs as it may get trapped in conditional continuous optima that differ from the global optimum.

The integrated \oneoneEAES with reduced success rule solves mixed discrete-continuous problems by creating one discrete and one continuous offspring in each iteration by using the mutations of the \oneoneEA and \oneoneES. It always accepts either both or neither of the offspring depending on whether they decrease or increase the joint objective.
Let $n_d, n_c \geq 1$ be the numbers of discrete and continuous variables, and $p_0 = \Brackets{1-1/n_d}^{n_d} \approx 1/e$ the probability that the discrete mutation does not flip any bit.
To prevent premature convergence, the algorithm updates the continuous step size using the $p_0/5$-rule rather than the usual $1/5$-rule.
With the success factor~${1 < \alpha < 1.4}$, the update mechanism works as follows.
Let $t \in \natnum$ denote the current iteration.
After the evaluation of the generated offspring, the algorithm increases the step size to $\sigma_{t+1}=\alpha \sigma_t$ when the joint offspring reduces the combined objective and decreases it otherwise to $\sigma_{t+1} = \sigma_t \alpha^{-p_0/\Brackets{5-p_0}}$.
To distinguish this algorithm from the \oneoneEAESBL, as introduced in \Cref{sect:algorithm}, we refer to the integrated \oneoneEAES by \oneoneEAESHB in the following.

The following lower bound identifies a trapping regime of the \oneoneEAESHB on \WPMCProximity.
Our subsequent analysis shows that the \oneoneEAESBL avoids this difficulty under its stated initialization and budget rule.
The experiments investigate how frequently the integrated algorithm enters the analyzed regime from random initialization.

We prove this result for the \WPMCProximity benchmark function, as defined in \Cref{sect:problem-desc}, to show that the \oneoneEAESHB can already require exponential expected time on mixed discrete-continuous problems with rather easy interactions that connect only pairs of variables.

As discussed in \Cref{sect:problem-desc}, the difficulty of the \WPMCProximity function is controlled by the parameter $w$ of the function, and for $0 <w < 1$, the function contains optimality gaps for all pairwise connected continuous and discrete variables.
Thus, if the algorithm reaches during its optimization process the $\epsilon$-range around a conditional continuous optimum, in which the continuous values are optimal for the current discrete bit string that differs from the optimal bit string, further discrete progress requires a sufficiently large accompanying change in the continuous variables to overcome the transition barrier.
Escaping this situation is especially unlikely for the algorithm when the step size of the continuous \oneoneES is so small that a sufficiently large continuous change to enable discrete progress is highly unlikely.
Near a conditional optimum, increasing the step size requires accepted offspring, while larger step sizes make continuous
improvements unlikely.
The proof below controls both accepted discrete changes and step-size increases that could permit escape.

To examine this intuition, we combine a theoretical analysis with an experimental investigation.
For $w < 1$, we define a trap as a state of the \oneoneEAESHB in which the continuous variables are close to a conditional continuous optimum and the step size is sufficiently small.
Specifically, let
\[
    \epsilon = \minWBrackets{\frac{w}{2},\frac{\Brackets{1-\sqrt{w}}^2}{256}}
    \quad\text{and}\quad
    \tilde{\sigma}_N =
    \frac{1-\sqrt{\epsilon}-\sqrt{w+\epsilon}}
         {\sqrt{2(N+1)\ln 2}}.
\]
The algorithm is in a trap if
$\NormTwo{y-(1-x)} \leq \epsilon$ with $x\neq 1^N$ and
$\sigma \in (0,\tilde{\sigma}_N]$.
These conditions describe a neighborhood of one of the conditional continuous optima $y=1-x$ in which the small step size makes escape difficult.
Our theoretical analysis establishes that escaping such a trap requires exponential time.
The experiments complement this result by showing frequent trap entries under the specified random initialization, even in small dimensions.

We first investigate experimentally how frequently the algorithm enters a trap from random initialization.
To this end, we perform~$100$ independent runs for each combination of variable pairs $N \in \{2,5,10,20,40,60,80,100\}$ and weights $w \in \{0.1,0.3,0.5,0.7,0.9\}$.
We sample $x_0$ uniformly from $\{0,1\}^N$ and, independently,
$y_0$ uniformly from $[0,1]^N$, and set ${\sigma_0=1}$ and $\alpha=1.3$.
Each run receives a budget of $500N$ iterations.
For each weight, we choose
$\epsilon=\minWBrackets{w/2,\Brackets{1-\sqrt{w}}^2/256}$ and use the
corresponding threshold $\tilde{\sigma}_N$ defined above.
We record a trap entry if, at some observed iteration $t\in \natnum$,
$x_t \neq 1^N$, $\NormTwo{y_t-\Brackets{1-x_t}} \leq \epsilon$, and
$\sigma_t \leq \tilde{\sigma}_N$ hold simultaneously.

\begin{figure}[t]
    \centering
    \includegraphics[width=\linewidth]
        {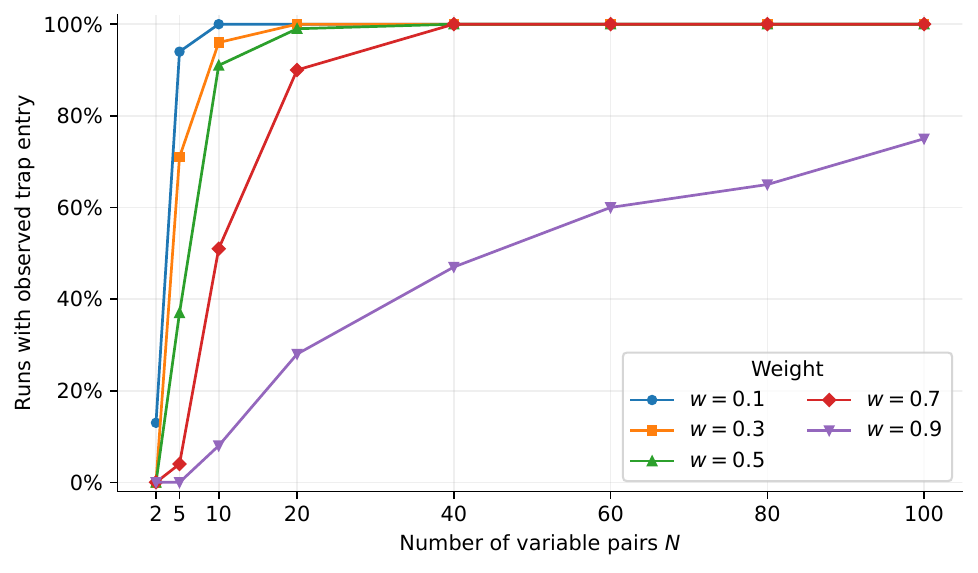}
    \caption{Observed trap-entry frequencies of the
        \oneoneEAESHB on \WPMCProximity,
        using $100$ independent runs per $(N,w)$ and a budget of
        $500N$ iterations per run. Each line represents one weight.}
    \Description{Line plot of observed trap-entry frequencies against
    the number of variable pairs N. Five curves show weights
    0.1, 0.3, 0.5, 0.7, and 0.9. Frequencies increase with N and
    are lower for larger weights under the tested settings.}
    \label{fig:integrated-trap-grid}
\end{figure}

\Cref{fig:integrated-trap-grid} shows that the observed entry
frequency increases with~$N$ and decreases with~$w$ under these
settings. Already for $N=10$, the entry frequencies are
$100\%$, $96\%$, and $91\%$ for~$w=0.1$,~$0.3$, and~$0.5$,
respectively. For every tested $N \geq 40$, all $100$ runs enter
the trap for each tested $w \leq 0.7$.
For $w=0.9$, an entry is less frequent, increasing from $8\%$ at
$N=10$ to $75\%$ at $N=100$, which is consistent with less frequent trapping at larger weights under the corresponding trap thresholds.
These finite-budget observations show frequent entry into the analyzed trap states under the tested settings.
They do not establish an asymptotic entry probability from random initialization.

These observations motivate the following theoretical analysis, which establishes an exponential lower bound on the expected optimization time when the algorithm is initialized in a trap.

\begin{theorem}
    \label{thm:lb-oneoneEAESHB-WPMCProximity}
    Let $n_d = n_c = N \geq2$ be the number of continuous and discrete variables.
    Consider the \oneoneEAESHB with success factor $1<\alpha<1.4$ optimizing \WPMCProximity as fitness function~$f$ with weight $w \in (0,1)$.
    Choose a fixed constant $\epsilon \in \left (0,\minWBrackets{w/2,\Brackets{1-\sqrt{w}}^2/256}\right ]$ and define
    $\delta:= 1-\sqrt{\epsilon} - \sqrt{w+\epsilon}$, and $\tilde{\sigma}_N:= \delta/(\sqrt{2(N+1)\ln 2})$.
    Let $x_0\in\{0,1\}^{n_d}$ be sampled uniformly at random.
    Initialize the continuous variables and the step size such that $\NormTwo{y_0-\Brackets{1-x_0}}\leq \epsilon$ and $\sigma_0 \in \left(0,\tilde{\sigma}_N\right]$.
    For every integer $t\geq0$, let $\theta_t=\Brackets{x_t,y_t,\sigma_t}$ denote the state after $t$ offspring evaluations, and define $T_\epsilon:=\inf\{t\geq0:f\Brackets{x_t,y_t}\leq\epsilon\}$.
    Then,
    \begin{align*}
        \Ex{T_\epsilon} &\geq 2^N.
    \end{align*}
\end{theorem}

\begin{proof}[Proof sketch]
        Conditional on the initial discrete bit string $x_0$ not being the optimal all-$1$ bit string, the algorithm can only leave the trap by either accepting an offspring with a changed discrete component or by increasing the step size above $\tilde{\sigma}_N$.
    Until the first accepted discrete change, elitism of the algorithm ensures that $\NormTwo{y_t-\Brackets{1-x_0}}\leq\epsilon$.
    Thus, continuous improvements alone cannot leave the trap, and reaching the target requires an accepted discrete change.

    For the first event, we show that the probability of accepting an offspring with a changed discrete component is exponentially small in $N$, since acceptance requires at least one flipped zero bit whose corresponding continuous coordinate decreases by more than $\delta$.
    A Gaussian tail bound using that the step size is smaller than $\tilde{\sigma}_N$ and a union bound over the initially zero bits give a probability of at most~$2^{-(N+1)}$.

    For the second event, crossing the step-size threshold without an accepted discrete change requires a continuous improvement at $\sigma_t>\tilde{\sigma}_N/\alpha$.
    Bounding the Gaussian density by its maximum and integrating over the improving ball of radius at most $\sqrt{\epsilon}$ gives a
    probability of at most $4^{-N}$.

    Both bounds hold conditional on every history before either event occurs.
    The expected waiting time until the first such event is therefore at least $\Brackets{2^{-(N+1)}+4^{-N}}^{-1}$.
    Since the target cannot be reached earlier and  $\Prob{x_0\neq 1^N}=1-2^{-N}$, we obtain
    \begingroup
    \setlength{\abovedisplayskip}{3pt}
    \setlength{\abovedisplayshortskip}{0pt}
    \setlength{\belowdisplayskip}{0pt}
    \setlength{\belowdisplayshortskip}{0pt}
    \[
        \Ex{T_\epsilon}
        \geq \frac{1-2^{-N}}{2^{-(N+1)}+4^{-N}}
        \geq 2^N.
    \]
    \endgroup
\end{proof}

\begin{proof}
    We bound the probability that either an offspring with a changed discrete component is accepted or the step size exceeds~$\tilde{\sigma}_N$.
    Continuous improvements that leave the discrete component unchanged are allowed throughout the argument.

    Using $\sqrt{w+\epsilon}\leq\sqrt{w}+\sqrt{\epsilon}$ and the upper bound on $\epsilon$, we obtain
    \begin{align*}
        \delta = 1-\sqrt{\epsilon} - \sqrt{w+\epsilon} \geq 1-\sqrt{w} - 2\sqrt{\epsilon} \geq 14\sqrt{\epsilon} >0.
    \end{align*}

    Consider the event~$A:= \CurlyBrackets{x_0 \neq 1^N}$, which describes that the initial discrete bit string is not the optimal all-$1$ bit string.
    Since~$x_0$ is sampled uniformly at random, $\Prob{A} = 1-2^{-N}$.

    For the following reasoning, we condition on $A$, and denote the number of initially zero bits by $k := \sum_{j=1}^{N} \Brackets{1-x_{0,j}}\geq 1$.

    Define the stopping times
    \begin{align*}
        \tau &:= \inf\CurlyBrackets{t\geq 1\mid x_t\neq x_0},\\\upsilon &:= \inf\CurlyBrackets{t\geq 1\mid \sigma_t> \tilde{\sigma}_N},\\ \rho &:= \minWBrackets{\tau,\upsilon}.
    \end{align*}
    Before $\tau$, the discrete contribution to the joint objective remains~$wk$.
    By elitism, the continuous error therefore satisfies for every $t<\tau$ that
    \begin{align*}
        \NormTwo{y_t-\Brackets{1-x_0}} \leq \NormTwo{y_0-\Brackets{1-x_0}} \leq \epsilon.
    \end{align*}
    Moreover, for $t <\tau$ it holds that $f\Brackets{x_t,y_t} \geq wk \geq w > \epsilon$, and hence
    \begin{align*}
        T_\epsilon \geq \tau \geq \rho.
    \end{align*}
    Now consider any iteration $t < \rho$.
    Write
    \begin{align*}
        z_t := y_t - \Brackets{1-x_0}, \quad r_t := \NormEucl{z_t} \leq \sqrt{\epsilon},
    \end{align*}
    and let $Z_t \sim \Normal\Brackets{0, I_N}$ denote the Gaussian mutation vector.
    All one-step probability bounds below hold conditional on the history up to iteration $t$.

    We first consider a step-size increase without an accepted discrete change.
    A rejected offspring decreases the step size, whereas an accepted offspring multiplies it by $\alpha$.
    Consequently, exceeding $\tilde{\sigma}_N$ requires $\sigma_t > \tilde{\sigma}_N/\alpha$.
    It therefore suffices to bound the probability of a continuous success at these step sizes.
    Successes at smaller step sizes are allowed, but cannot immediately raise the step size above~$\tilde{\sigma}_N$.

    If there are no bit flips, the discrete contribution to the objective remains unchanged.
    Hence, the offspring is accepted precisely when its distance to the conditional continuous optimum decreases, that is, when
    \begin{align*}
        \NormEucl{z_t + \sigma_tZ_t} < r_t.
    \end{align*}
    Thus, the successful offspring lies in the ball
    \begin{align*}
        B_N\Brackets{0,r_t}:= \CurlyBrackets{u \in \realnum^N: \NormEucl{u} < r_t}.
    \end{align*}
    Conditional on the current state, the continuous offspring relative to the conditional optimum follows the normal distribution $\Normal\Brackets{z_t,\sigma_t^2I_N}$.
    Its density is bounded everywhere by $\Brackets{2\pi\sigma_t^2}^{-N/2}$.
    Integrating this density over the successful region, and using
    \begin{align*}
        \text{vol}\Brackets{B_N(0,r)}\leq \Brackets{\frac{2\pi e r^2}{N}}^{N/2},
    \end{align*}
    we obtain
    \begin{align*}
        \Prob{\NormEucl{z_t + \sigma_tZ_t}< r_t \mid \theta_t} &= \int_{B_N\Brackets{0,r_t}}\frac{\exp\Brackets{-\frac{\NormTwo{u-z_t}}{2\sigma_t^2}}}{\Brackets{2\pi \sigma_t^2}^{N/2}}du\\
        &\leq \Brackets{2\pi \sigma_t^2}^{-N/2}\text{vol}\Brackets{B_N\Brackets{0,r_t}}\\
        &\leq \Brackets{\frac{er_t^2}{N\sigma_t^2}}^{N/2}.
    \end{align*}
    Thus, we upper-bound the success probability by the maximal Gaussian density multiplied by the volume of the successful region.

    We now use the largest possible radius, $r_t \leq \sqrt{\epsilon}$, and the smallest step size that can lead to a crossing, $\sigma_t > \tilde{\sigma}_N/\alpha$.
    In the following calculation, we insert those values in the previous upper bound for $r_t$ and $\sigma_t$, and use $(N+1)/N\leq 3/2$ in the second inequality.
    Using $\delta^2\geq 196\epsilon$ and $\alpha < 1.4$ for the third inequality and ${3e\ln 2/100 < 1/16}$ for the last, we get
    \begin{align*}
        \Prob{\NormEucl{z_t + \sigma_tZ_t}< r_t \mid \theta_t} &\leq \Brackets{\frac{2e\alpha^2\epsilon(N+1)\ln2}{N\delta^2}}^{N/2}\\
        &\leq \Brackets{\frac{3e\alpha^2\epsilon\ln2}{\delta^2}}^{N/2}\\
        &\leq \Brackets{\frac{3e\ln2}{100}}^{N/2}\\
        &\leq 4^{-N}.
    \end{align*}
    Thus, the choice of $\epsilon$ ensures that, at any step size permitting a crossing of $\tilde{\sigma}_N$, the probability of sampling an improving continuous offspring is exponentially small in $N$.
    An actual continuous success additionally requires that no bit flips.
    Since the discrete and continuous mutations are independent, imposing this condition can only decrease the probability.
    Therefore, for every history before~$\rho$, the probability that a continuous success raises the step size above~$\tilde{\sigma}_N$ in the next iteration is at most $4^{-N}$.

    Next, we bound the probability of accepting an offspring with a changed discrete component.
    Let $a$ and $b$ denote the numbers of bits flipped from $0$ to $1$ and from $1$ to $0$, respectively.
    If $a=0$ and $b\geq 1$, the discrete contribution increases by at least $w>\epsilon$, which cannot be compensated by a continuous improvement.
    Hence, acceptance of a changed bit string requires $a \geq 1$.

    For an accepted offspring $\Brackets{x_t',y_t'}$, we have
    \begin{align*}
        f\Brackets{x_t',y_t'} = w(k-a+b) + \NormTwo{y_t'-\Brackets{1-x_t'}} < wk + r_t^2,
    \end{align*}
    and therefore
    \begin{align*}
        \NormTwo{y_t'-\Brackets{1-x_t'}} < w(a-b) + r_t^2 \leq wa + \epsilon.
    \end{align*}
    Consequently, at least one index $j$ with a $0$-to-$1$ flip must satisfy
    \begin{align}
        \Brackets{y_{t,j}'}^2 < w + \epsilon \label{eq:lb:integrated:cond}.
    \end{align}

    Let $Z_t = \Brackets{Z_{t,1},...,Z_{t,N}} \sim \Normal\Brackets{0,I_N}$ denote the Gaussian vector used for the continuous mutation.
    Since $t<\tau$, elitism gives $\NormTwo{y_t-\Brackets{1-x_0}}\leq\epsilon$.
    Thus, $y_{t,j}\geq1-\sqrt{\epsilon}$ for every initially zero bit.
    Using $y_{t,j}' = y_{t,j} + \sigma_t Z_{t,j}$, the condition in \Cref{eq:lb:integrated:cond} implies
    \begin{align*}
        &\Brackets{y_{t,j}+\sigma_tZ_{t,j}}^2 < w+\epsilon \Leftrightarrow \lvert y_{t,j}+\sigma_tZ_{t,j}\rvert < \sqrt{w+\epsilon} \\&\Rightarrow 1 - \sqrt{\epsilon}+\sigma_tZ_{t,j} < \sqrt{w+\epsilon},
    \end{align*}
    and therefore
    \begin{align*}
        Z_{t,j} < - \frac{1-\sqrt{\epsilon}-\sqrt{w+\epsilon}}{\sigma_t} = - \frac{\delta}{\sigma_t}.
    \end{align*}

    Let $\Phi_1$ denote the standard normal cumulative distribution function.
    In the following, we use the independence of the discrete and continuous mutations, a union bound over the $k$ initially zero bits, and the Gaussian tail bound.
    With $\sigma_t \leq \tilde{\sigma}_N$, we obtain
    \begin{align*}
        \Prob{x_{t+1}\neq x_t \mid \theta_t} &\leq \frac{k}{N} \Phi_1\Brackets{-\frac{\delta}{\sigma_t}}\\
        &\leq \exp\Brackets{-\frac{\delta^2}{2\tilde{\sigma}_N^2}}\\
        &\leq 2^{-(N+1)}.
    \end{align*}

    The stopping time $\rho$ can be reached only through an accepted discrete change or through a continuous success that raises the step size above $\tilde{\sigma}_N$.
    Combining the two bounds yields
    \begin{align*}
        \Prob{\rho = t+1\mid \rho > t, A} \leq 2^{-(N+1)} +4^{-N} =: q_N.
    \end{align*}
    Since this estimate holds for every history before $\rho$, it follows inductively that
    \begin{align*}
        \Prob{\rho > t\mid A} \geq \Brackets{1-q_N}^t.
    \end{align*}
    Using the tail-sum formula for non-negative integer-valued random variables, we obtain
    \begin{align*}
        \Ex{\rho \mid A} &= \sum_{t=0}^\infty \Prob{\rho > t \mid A} \geq \sum_{t=0}^\infty \Brackets{1-q_N}^t\\
        &= 1/q_N.
    \end{align*}
    Finally, since $T_\epsilon \geq \rho$ and using the law of total expectation and $N\geq 2$, we get
    \begin{align*}
        \Ex{T_\epsilon} &\geq \Prob{A} \Ex{\rho\mid A}\\
        &\geq \frac{1-2^{-N}}{2^{-(N+1)}+4^{-N}}\\
        &= \frac{2^{N+1}\Brackets{2^N-1}}{2^N+2} \geq 2^N.
    \end{align*}
    The conditional bound $\Ex{T_\epsilon\mid\theta_0}\geq1/q_N$ also holds for every fixed initial trap state, since all one-step estimates are uniform over such states.
\end{proof}
\section{Definition of \oneoneEAESBL}
\label{sect:algorithm}

In order to solve mixed discrete-continuous problems with interactions between the discrete and continuous parts, we define the \oneoneEAESBL as follows.
At the upper level, the algorithm mutates the discrete part of the current best solution in each iteration and then calls the lower-level continuous algorithm, which optimizes the continuous part while keeping the discrete offspring fixed.
After a fixed budget of fitness evaluations, the lower-level algorithm returns the continuous offspring, and the algorithm replaces the current best solution if the joint offspring has a mixed fitness value no larger than the current one.

We use the mutation operator of the standard \oneoneEA, which flips each bit of the discrete bit string independently with probability $1/n_d$.
The \oneoneES on the lower level starts each call at the same continuous point $y_0$ with step size~$1$; we denote this initialization by \textsc{InitGen}.
We retain the incumbent joint solution and its exact fitness value; obtaining a new approximation of the conditionally minimized objective would require another lower-level run.
Retaining previously obtained fitness values has also been studied in noisy evolutionary optimization by Antipov and Doerr~\cite{DoerrNoise2025}.
Here, we analyze this choice for the residual errors induced by the lower-level optimizer.

With $\textsc{StopCritCap}$ we denote the stopping criterion which fixes the budget of fitness evaluations for each lower-level call.
Let $C>0$ be a constant, $R\geq e$ an upper bound on the distances from $y_0$ to all relevant conditional optima, and $\epsilon\in(0,1)$.
Then the budget is given by the function
\begin{align}\label{eq:bilevel-alg:ccap}
    c_{\text{cap}}\Brackets{n_c,R} = \left\lceil C\cdot n_c\cdot \log \Brackets{R/\sqrt{\epsilon}}\right\rceil.
\end{align}
It includes the evaluation of the initial continuous point, followed by $c_{\text{cap}}\Brackets{n_c,R}-1$ mutations.
As shown in the next section in \Cref{lem:akimoto-lower-bound-prob}, for sufficiently large $C$ and any valid distance bound $R$, each call returns a continuous residual error of at most $\epsilon$ with probability at least $1- \BigO{1/C}$.

\Cref{algo:one-one-ea-es-bl,algo:one-one-ea-es-bl-lower-level} give the upper and lower levels of the algorithm with $B=c_{\text{cap}}\Brackets{n_c,R}$.

\begin{algorithm}[htp]
\caption{The \oneoneEAESBL with fixed continuous initialization and lower-level budget.}\label{algo:one-one-ea-es-bl}
\KwIn{fitness function $f\colon\{0,1\}^{n_d}\times\realnum^{n_c}\to\realnum$; fixed starting point $y_0$; success factor $\alpha>1$; budget $B=c_{\text{cap}}\Brackets{n_c,R}$\;}
Sample $x$ uniformly from $\{0,1\}^{n_d}$ independently of $y_0$\;
$y\leftarrow y_0$ and $F\leftarrow f(x,y)$\;
\While{overall stopping criterion not met}{
    $x' \leftarrow \text{discrete-mutation}_{1/n_d}(x)$\;
    $(F',y') \leftarrow \textsc{ContAlg}\Brackets{f(x',\cdot),y_0,\alpha,B}$\;
    \If{$F'\leq F$}{
        $(x,y,F)\leftarrow(x',y',F')$\;
    }
}
\Return{$(x,y)$}\;
\end{algorithm}

\begin{algorithm}[htp]
\caption{The lower-level \textsc{ContAlg} with fixed discrete offspring $x'$.}\label{algo:one-one-ea-es-bl-lower-level}
\KwIn{fitness function $f(x',\cdot)$; starting point $y_0$; success factor $\alpha>1$; budget $B$\;}
$z\leftarrow y_0$, $\sigma\leftarrow 1$, and $F\leftarrow f(x',z)$\;
\For{$s=1,\ldots,B-1$}{
    $z'\leftarrow z+\sigma\Normal\Brackets{0,I_{n_c}}$\;
    $F'\leftarrow f(x',z')$\;
    \uIf{$F'\leq F$}{
        $(z,F)\leftarrow(z',F')$\; $\sigma\leftarrow\sigma\alpha$\;
    }
    \Else{
        $\sigma\leftarrow\sigma\alpha^{-1/4}$\;
    }
}
\Return{$(F,z)$}\;
\end{algorithm}
\section{Fixed-Budget Guarantees for the Lower-Level Optimizer}
\label{sect:fixed-budget-guarantees}

Building on the result of Akimoto et al.\ in \cite{akimotoDriftTheoryContSearchSpaces2018}, we derive a fixed-budget success guarantee for the \oneoneES on the \Sphere function, which holds uniformly over all relevant dimensions and admissible initial points.
This results in a guarantee for the lower-level optimizer of the \oneoneEAESBL when applied on a problem with a \Sphere-shaped continuous part as in all benchmark problems defined in \Cref{sect:problem-desc}.

In the following, we denote, for $d \in \natnum$ and $z \in \realnum^{d}$, the $d$-dimensional \Sphere function by $h_d(z):= \NormTwo{z}$.
For ${q>0}$ and with $\log^+(z):= \maxWBrackets{0,\log z}$, we define
\(
L(q) := \log R + 1/2 \log^+(1/q).
\)

\begin{lemma}
\label{lem:akimoto-lower-bound-prob}
    Let $n_c \geq 1$ and $d\in \natnum$ with $d \leq n_c$.
    Consider the \oneoneES with fixed success factor $\alpha > 1$ and the initial step size $\sigma_{\text{init}} >0$ optimizing the $d$-dimensional \Sphere function $h_d$.
    There exists a constant $M>0$, depending only on $\alpha$ and $\sigma_{\text{init}}$, such that the following holds.
    Let $z_0\in\realnum^d$ be the initial point with $R\geq\maxWBrackets{e,\NormEucl{z_0}}$. Assume $\epsilon\in(0,1)$.
    With $L$ as defined above, for a constant $C \geq 2$, let $c_{\text{cap}}\Brackets{n_c,R} := \left\lceil Cn_cL(\epsilon)\right\rceil$ be the number of fitness evaluations, and let~$Z_{\text{out}}$ denote the returned point after those.
    Then, for every $q > 0$, the probability that the returned point has a value of at most $q$, is lower-bounded by
    \begin{align*}
        \Prob{h_d\Brackets{Z_{\text{out}}} \leq q} \geq p_{\leq q} := \maxWBrackets{0, 1-\frac{M}{C}\frac{L(q)}{L(\epsilon)}}.
    \end{align*}
    In particular, for $q = \epsilon$ it applies that
    \begin{align*}
        \Prob{h_d\Brackets{Z_{\text{out}}} \leq \epsilon} \geq p_{\leq \epsilon} = \maxWBrackets{0, 1-\frac{M}{C}}.
    \end{align*}
    Moreover, $p_{\leq q}$ is non-decreasing in $q$.
\end{lemma}

\begin{proof}[Proof sketch]
    We obtain the result by first using the potential function and parameter choices of Akimoto et al.\ in \cite{akimotoDriftTheoryContSearchSpaces2018}.
    This potential combines the logarithmic distance to the optimum with a penalty for poorly adapted step sizes.
    We lower-bound its truncated drift by a positive constant times $1/d$, uniformly over $d\geq1$.
    Next, we show that the penalty term for the initial step size is at most a constant multiple of $L(q)$, uniformly over all dimensions and admissible starting points.
    Thus, the initial potential lies at most~$\BigO{L(q)}$ above the target level.
    With $\tau_q$ counting the mutations until $h_d\leq q$, there is a~$K_0$ depending only on the fixed algorithm parameters, for which the truncated drift theorem yields
    \[
        \Ex{\tau_q}\leq K_0dL(q).
    \]
    The budget permits at least $(C/2)n_cL(\epsilon)$ mutations.
    By elitism and Markov's inequality, the failure probability is therefore at most
    \[
        \frac{2K_0}{C}\frac{d}{n_c}\frac{L(q)}{L(\epsilon)}.
    \]
    Choosing $M:= 2K_0$ proves the probability bound.
    For $d=0$, the error is identically zero.
    Since $L$ is non-increasing, $p_{\leq q}$ is non-decreasing in $q$, which completes the proof.
\end{proof}

For the subsequent upper-level analysis, uniformity is essential: the discrete offspring depends on previous selections.
The guarantee nevertheless applies conditional on every preceding history, so a fixed choice $C>M$ gives the same positive lower bound on success in each call.

\begin{proof}
    We use the \href{https://arxiv.org/abs/1802.03209v4}{corrected arXiv version} of \cite{akimotoDriftTheoryContSearchSpaces2018}.
    Result numbers refer to the conference version.

    We first establish a dimension-uniform drift bound and then control the initial potential uniformly over all admissible starting points.

    Let $l_d,u_d$ be admissible normalized step-size bounds and $v_d>0$ a weight to be chosen below.
    For $d \geq 1$, consider the potential from Akimoto et al.\ in \cite[Equation~8]{akimotoDriftTheoryContSearchSpaces2018},
    \begin{align*}
        V_d(z,\sigma) = \log\NormEucl{z} + v_d \maxWBrackets{0,\log\frac{\alpha l_d \NormEucl{z}}{d\sigma}, \log\frac{\alpha^{1/4}d\sigma}{u_d\NormEucl{z}}}.
    \end{align*}
    Let $A_d$ denote the truncation parameter and $B_d$ the drift bound in~\cite[Proposition~4.2]{akimotoDriftTheoryContSearchSpaces2018}.

    Fix an integer $D\geq2$ such that $D\log\alpha>1$.
    For $d\geq D$, the parameter choices of \cite[Proposition~4.3]{akimotoDriftTheoryContSearchSpaces2018} with $A_d=1/d$ yield $B_d>0$ and $B_d\in\Theta(1/d)$.

    For each remaining dimension $1\leq d<D$, set $A_d=1/d$ and apply the general truncated drift bound.
    As $v_d\downarrow0$, the required relative improvement converges to $1-e^{-1/d}$.
    By continuity and compactness of the admissible step-size interval, the corresponding minimum success probability stays bounded
    away from zero.
    Choosing $v_d \in \Brackets{0,\minWBrackets{1,A_d/\log\alpha}}$ sufficiently small therefore makes all terms in the minimum defining $B_d$ positive.

    Combining these finitely many choices with the asymptotic bound gives
    \[
        \beta:=\inf_{d\geq1} dB_d>0.
    \]

    We now bound the initial potential uniformly over all dimensions and admissible starting points.
    The parameter choice in \cite[Proposition~4.3]{akimotoDriftTheoryContSearchSpaces2018} gives $v_d = \BigO{1/d}$.
    The normalized step-size thresholds $l_d,u_d$ are chosen using fixed success probabilities; by \cite[Lemmas~3.1~and~3.2]{akimotoDriftTheoryContSearchSpaces2018}, these thresholds converge to positive finite limits.
    Enlarging the constants to cover the finitely many remaining dimensions, including the previously fixed choice for $d=1$, therefore yields a constant $c_0 \geq 1$ such that
    \begin{align*}
        v_d \leq \frac{c_0}{d},\quad l_d \leq c_0, \quad u_d^{-1} \leq c_0
    \end{align*}
    for every $d \geq 1$.

    Fix $q > 0$.
    If $\NormTwo{z_0}\leq q$, the claim of the lemma follows from elitism of the \oneoneES.
    Otherwise, write $r := \NormEucl{z_0} > \sqrt{q}$ and define
    \begin{align*}
        \kappa_d := V_d\Brackets{z_0,\sigma_{\text{init}}} - \log r.
    \end{align*}
    By the definition of $V_d$,
    \begin{align*}
        \kappa_d = v_d \maxWBrackets{0,\log\frac{\alpha l_d r}{d\sigma_{\text{init}}}, \log\frac{\alpha^{1/4}d\sigma_{\text{init}}}{u_d r}}.
    \end{align*}
    The first logarithmic term increases with $r$ and the second decreases; using $\sqrt{q}<r\leq R$ and $L(q)\geq1$, we obtain a constant $c_1>0$, depending only on $\alpha$ and $\sigma_{\text{init}}$, such that
    \begin{align*}
        \kappa_d\leq\frac{c_1}{d}\Brackets{L(q)+\log d}.
    \end{align*}

    Let $z_t$ denote the incumbent after $t$ mutations of a run without a budget limit, and define $\tau_q:=\inf\CurlyBrackets{t\in\natnum\mid h_d(z_t)\leq q}$ as the number of mutations until the target $q$ is reached.
    Since $V_d(z,\sigma) \geq \log \NormEucl{z}$, reaching $V_d \leq \log \sqrt{q}$ implies $\NormTwo{z}\leq q$.
    Thus, the corresponding hitting time of the potential upper-bounds $\tau_q$.
    Applying the truncated drift theorem, as given in \cite[Theorem~2.1]{akimotoDriftTheoryContSearchSpaces2018}, to $V_d$, with $A_d=1/d$ and $B_d\geq \beta/d$, gives
    \begin{align*}
        \Ex{\tau_q} &\leq \frac{V_d\Brackets{z_0,\sigma_{\text{init}}} - \log\sqrt{q} +1/d}{B_d}\\
        &\leq \frac{d}{\beta} \Brackets{\log \frac{r}{\sqrt{q}} + \kappa_d +\frac{1}{d}}\\
        &\leq \frac{1}{\beta} \Brackets{dL(q)+c_1\Brackets{L(q)+\log d}+1}.
    \end{align*}
    Since $L(q) \geq 1$, $d\geq 1$, and $\log d \leq d$, choosing the constant
    \begin{align*}
        K_0:= \frac{2+2c_1}{\beta}
    \end{align*}
    gives $\Ex{\tau_q} \leq K_0 dL(q)$.

    With the given budget $c_{\text{cap}}\Brackets{n_c,R}$, the \oneoneES performs $t_{\text{cap}}:= c_{\text{cap}}\Brackets{n_c,R}-1$ mutations.
    Since $C\geq 2$, $n_c\geq 1$, and $L(\epsilon)\geq 1$, it holds that
    \begin{align*}
        t_{\text{cap}} \geq Cn_cL(\epsilon) -1 \geq \frac{C}{2}n_cL(\epsilon).
    \end{align*}
    By elitism and Markov's inequality, we get
    \begin{align*}
        \Prob{h_d\Brackets{Z_{\text{out}}}>q} &= \Prob{\tau_q >t_{\text{cap}}}\\
        &\leq \frac{K_0dL(q)}{t_{\text{cap}}}\\
        &\leq \frac{2K_0}{C}\frac{d}{n_c}\frac{L(q)}{L(\epsilon)}\\
        &\leq \frac{2K_0}{C}\frac{L(q)}{L(\epsilon)}.
    \end{align*}
    Choosing $M:= 2K_0$ proves the probability bound.
    For $d=0$, the error is identically zero.
    Since $L$ is nonincreasing, $p_{\leq q}$ is non-decreasing in $q$, which completes the proof.
\end{proof}
\section{Run Time Analysis of the \oneoneEAESBL}
\label{sect:analysis-bilevel}

By \Cref{tab:functions:gen-sphere-lin}, \SphereOneMax, Type-II, and \WPMCProximity belong to the \GenSphereLinear function class.
In the following, we first establish lower and upper bounds on the expected first hitting time of the \oneoneEAESBL for this class and then specialize them to the individual benchmark functions.

We analyze the \oneoneEAESBL using the general initialization \textsc{InitGen} and the maximum budget stopping criterion \textsc{StopCritCap} for the continuous lower-level algorithm, as defined in \Cref{sect:algorithm}.
Throughout this section, we condition on an arbitrary fixed
continuous initial point $y_0 \in \realnum^{n_c}$.
The discrete initial point $x_0$ is sampled uniformly from
$\{0,1\}^{n_d}$, independently of the continuous initialization.
All expectations below are conditional on $y_0$.
Consequently, the distance bound and the lower-level budget
are deterministic.
Choose an available distance bound $R_{\text{GPOM}}$ of a function of the class \GenSphereLinear satisfying
\begin{align}\label{eq:bilevel:benchmark-proofs:def-R}
R_{\text{GPOM}}
\geq \maxWBrackets{e,\max_{x\in\{0,1\}^{n_d}}
\NormEucl{y_0-y^*(x)}}.
\end{align}
 With a constant $C>0$, we use for the stopping criterion \textsc{StopCritCap} the budget $c_{\text{cap}}\Brackets{n_c,R_{\text{GPOM}}} = \left\lceil C \cdot n_c \cdot \logWBrackets{R_{\text{GPOM}}/\sqrt{\epsilon}}\right\rceil$, as defined in \Cref{eq:bilevel-alg:ccap}.
It counts all fitness evaluations performed within a lower-level call, including the evaluation of its initial solution.

The next theorem derives a run time lower bound by applying Witt's comparison argument~\cite{WITT_2013} to the discrete generation process and accounting for the fixed lower-level budget.

\begin{theorem}
\label{thm:bilevel:gen:lb}
    Let $n_d \geq 2$ and $n_c \geq 1$ denote the number of discrete and continuous variables.
    Consider the \oneoneEAESBL optimizing a fitness function~$f$ of the class \GenSphereLinear with weight $w > 0$ and assume $\epsilon\in\Brackets{0,\minWBrackets{1,w}}$.
    With $R_{\text{GPOM}}$ chosen as in~\Cref{eq:bilevel:benchmark-proofs:def-R}, use the budget $c_{\text{cap}}\Brackets{n_c,R_{\text{GPOM}}}$ as stopping criterion of the lower-level algorithm.
    Let $T_\epsilon$ denote the number of fitness evaluations performed after initialization until the algorithm first evaluates a solution with an objective value at most $\epsilon$.
    Then, the expected first hitting time is bounded from below by
    \begin{align*}
        \Ex{T_\epsilon} &\geq (1-o(1))\cdot e \cdot n_d \cdot \ln\Brackets{n_d} \cdot c_{\text{cap}}\Brackets{n_c,R_{\text{GPOM}}}.
    \end{align*}
\end{theorem}

\begin{proof}[Proof sketch]
    In order to obtain the lower bound, we first analyze the number of upper-level iterations the algorithm requires in expectation until the discrete optimum is generated for the first time.
    To obtain this, we compare this process of the \oneoneEAESBL with the standard \oneoneEA on \OneMax.
    Since every discrete offspring is generated by standard bit mutation with probability $1/n_d$ and its parent is a previously generated bit string, the comparison argument of Witt in \cite[Theorem~6.2]{WITT_2013} can be applied.
    The key property is that, for mutation probability at most $1/2$, the offspring's Hamming distance from the optimum is stochastically non-decreasing in the parent's distance.
    This comparison remains valid conditional on the complete history, including all previous continuous computations.
    Thus, the first generation time of the discrete optimum stochastically dominates the corresponding time of the \oneoneEA on \OneMax and it is possible to apply the lower-bound of $(1-o(1))\cdot e \cdot n_d \cdot \ln\Brackets{n_d}$, as stated in \Cref{thm:witt-runtime-linear}.

    Since $\epsilon < w$, no lower-level call can reach an objective value at most $\epsilon$ before the first generation of the discrete optimum.
    Every lower-level call preceding the first generation of the discrete optimum uses the full budget.
    Allowing for a possibly incomplete call when the discrete optimum is first generated, we subtract one from the lower bound on the expected number of upper-level iterations before multiplying by $c_{\text{cap}}\Brackets{n_c,R_{\text{GPOM}}}$.
    This subtraction is absorbed into the $o(1)$ term.
\end{proof}

\begin{proof}
    In the following, we first lower-bound the number of upper-level iterations required to generate the discrete optimum by comparing the discrete generation process with the standard \oneoneEA on \OneMax.
    We then translate this bound into fitness evaluations using the fixed budget of each lower-level call.

    Set $X^{(0)}:= x_0$.
    For every integer $u \geq 1$, let $X^{(u)}$ denote the discrete offspring generated in the $u$-th upper-level iteration.
    Define
    \begin{align*}
        U := \inf \Brackets{u \in \natnum \mid X^{(u)} = 1^{n_d}}.
    \end{align*}
    Thus, $U$ counts the upper-level iterations until the discrete optimum is first generated.
    Acceptance of the generated bit string is not required in this definition.

    Since $\epsilon < w$, before the first generation of the discrete optimum, no lower-level call can reach an objective value at most $\epsilon$.
    If $U \geq 1$, the preceding $U-1$ calls have therefore been completed, each using $c_{\text{cap}}\Brackets{n_c,R_{\text{GPOM}}}$ fitness evaluations.
    With $\Brackets{z}_+ := \maxWBrackets{z,0}$, it follows that $T_\epsilon \geq c_{\text{cap}}\Brackets{n_c,R_{\text{GPOM}}}\cdot \Brackets{U-1}_+$.
    Taking expectations and using $\Brackets{U-1}_+\geq \Brackets{U-1}$, we obtain
    \begin{align}
        \Ex{T_\epsilon} \geq c_{\text{cap}}\Brackets{n_c,R_{\text{GPOM}}}\cdot \Brackets{\Ex{U}-1}.\label{eq:thm-lb-first-hit:lb}
    \end{align}
    It therefore remains to lower-bound $\Ex{U}$.

    The discrete generation process of the \oneoneEAESBL starts with a single uniformly sampled bit string.
    Every subsequent offspring is obtained by applying standard bit mutation with probability $p = 1/n_d \leq 1/2$ to a previously generated bit string.
    Conditional on the history before the next mutation, including all previous continuous computations, the selected parent is a previously generated bit string, and each bit is flipped independently with probability~$1/n_d$.
    The continuous optimization only affects the choice of the next parent.
    According to \cite[Theorem~6.2]{WITT_2013}, the optimization time of a mutation-based EA with population size $\mu=1$ and mutation probability $p \leq 1/2$ on any function with a unique global optimum is stochastically at least as large as the optimization time of the \oneoneEA on \OneMax.
    It is possible to use this theorem, as we apply the comparison argument only to the discrete generation process of the \oneoneEAESBL, which relies on discrete mutations.

    Let $U_{\text{OM}}$ denote the number of mutation steps required by the standard \oneoneEA, initialized uniformly at random, to first generate the optimum of $\text{OM}(x):= \sum_{i=1}^{n_d}\Brackets{1-x_i}$.

    The comparison argument in \cite[Theorem~6.2]{WITT_2013} is based on the monotonicity established in \cite[Lemma~6.1]{WITT_2013}: under bit mutations with $p \leq 1/2$, the offspring's Hamming distance from the target is stochastically non-decreasing in the parent's distance.
    The \OneMax process retains a previously generated search point with the minimum Hamming distance from the optimum, whereas every parent selected by the bilevel algorithm must belong to its previously generated search points.
    Consequently, for every integer~$u\geq 0$, we get
    \begin{align*}
        \Prob{U > u} \geq \Prob{U_{\text{OM}} > u},
    \end{align*}
    and hence $\Ex{U} \geq \Ex{U_{\text{OM}}}$.

    Applying \cite[Corollary~6.6]{WITT_2013}, as stated in \Cref{thm:witt-runtime-linear}, to the \OneMax comparison process with a single uniformly sampled initial point and mutation probability $1/n_d$ yields
    \begin{align*}
        \Ex{U} \geq \Ex{U_{\text{OM}}} \geq (1-o(1)) e n_d \ln n_d.
    \end{align*}
    Substituting this bound into \Cref{eq:thm-lb-first-hit:lb}, we obtain
    \begin{align*}
        \Ex{T_\epsilon} &\geq  c_{\text{cap}}\Brackets{n_c,R_{\text{GPOM}}}\cdot \Brackets{(1-o(1))\cdot e\cdot n_d \cdot \ln n_d-1}\\
        &= (1-o(1))\cdot e\cdot n_d\cdot \ln n_d\cdot c_{\text{cap}}\Brackets{n_c,R_{\text{GPOM}}},
    \end{align*}
    where $o(1)$ refers to $n_d \rightarrow \infty$.
    This completes the proof.
\end{proof}

In order to obtain the corresponding upper bound in the next theorem, we use the fixed-budget guarantee of \Cref{lem:akimoto-lower-bound-prob} and the multiplicative drift theorem, as stated in \Cref{thm:prelim:mult}.

\begin{theorem}
\label{thm:bilevel:gen:ub}
    Let $n_d \geq 2$ and $n_c \geq 1$ denote the number of discrete and continuous variables.
    Consider the \oneoneEAESBL optimizing a fitness function~$f$ of the class \GenSphereLinear with weight $w > 0$ and assume $\epsilon\in\Brackets{0,\minWBrackets{1,w}}$.
    Let $M$ be the constant from \Cref{lem:akimoto-lower-bound-prob} and choose a fixed constant $C\geq 2$ with $C > M$.
    With these constants and $R_{\text{GPOM}}$ chosen as in~\Cref{eq:bilevel:benchmark-proofs:def-R}, use the budget $c_{\text{cap}}\Brackets{n_c,R_{\text{GPOM}}}$ as stopping criterion of the lower-level algorithm.
    Let $T_\epsilon$ denote the number of fitness evaluations performed after initialization until the algorithm first evaluates a solution with an objective value at most $\epsilon$.
    Then, the expected first hitting time is bounded from above by
    \begin{align*}
        \Ex{T_\epsilon} &\leq (1+o(1))\cdot e \cdot n_d \cdot \ln\Brackets{n_d} \cdot \frac{C}{C-M} \cdot c_{\text{cap}}\Brackets{n_c,R_{\text{GPOM}}}.
    \end{align*}
\end{theorem}

\begin{proof}[Proof sketch]
    In order to obtain the upper bound, we first analyze the number of upper-level iterations the algorithm requires in expectation until the joint objective is at most $\epsilon$.

    For each integer $u \geq 0$, let $\Brackets{x_u,y_u}$ denote the incumbent after $u$ completed upper-level iterations and write $F_u:= f\Brackets{x_u,y_u}$.
    Define $U := \inf\CurlyBrackets{u \geq 0 \mid F_u \leq \epsilon}$, which counts the completed upper-level iterations until the incumbent first has objective value at most $\epsilon$.
    We measure the excess objective value in units of $w$ and cap the resulting potential at $n_d$.
    This controls the initial potential even when the continuous error is large.
    Thus, we define the potential process by
    \begin{align*}
        G_u := g\Brackets{x_u,y_u} = \minWBrackets{n_d, \left \lceil \frac{F_u-\epsilon}{w}\right\rceil}.
    \end{align*}
    It ensures $G_u = 0$ if and only if $F_u \leq \epsilon$, so that $U$ is precisely the first hitting time of zero by the potential process.
    Since the potential is non-decreasing in the retained objective value, elitist selection prevents it from increasing.

    \begingroup\emergencystretch=1em
    As established in \Cref{lem:akimoto-lower-bound-prob}, let $p_{\leq \epsilon}$ denote the lower bound on the probability that a lower-level call returns a continuous error of at most~$\epsilon$ uniformly over all previous histories and discrete offspring.\par\endgroup
    For $q_{n_d} := \frac{1}{n_d}\Brackets{1-1/n_d}^{n_d-1}$ as the probability that only a specific bit is flipped and with $\mathcal{F}_u$ as the complete history after $u$ completed upper-level iterations, we show that, whenever $u < U$, the conditional expected decrease satisfies
    \begin{align*}
        \Ex{G_u-G_{u+1} \mid \mathcal{F}_u} \geq q_{n_d} p_{\leq \epsilon} G_u.
    \end{align*}

    We obtain this by dividing the analysis into two cases according to whether the discrete part is optimal.

    Let $k$ denote the number of zero bits in $x_u$; then $G_u\geq k$.
    If $k\geq1$, flipping exactly one zero bit and obtaining a continuous error of at most $\epsilon$ produces an accepted offspring with potential $k-1$.
    Each of these $k$ disjoint events has conditional probability at least $q_{n_d}p_{\leq\epsilon}$ and decreases the potential by $G_u-k+1$.
    Since $k\Brackets{G_u-k+1}\geq G_u$, their contributions yield the claimed drift bound.

    \begingroup\emergencystretch=1em
    In the case that the discrete part is optimal, we show that the potential decreases to zero when the continuous error is at most $\epsilon$ and no bit is flipped.
    The probability of no bit flip is ${\Brackets{n_d-1}q_{n_d}\geq q_{n_d}}$.
    Thus, this event also yields an expected decrease of at least~$q_{n_d}p_{\leq\epsilon}G_u$.\par\endgroup

    Applying the multiplicative drift theorem with $g\Brackets{x_0,y_0} \leq n_d$, $q_{n_d}\geq1/\Brackets{en_d}$, and $p_{\leq \epsilon}\geq 1-M/C$ results in
    \begin{align*}
        \Ex{U} &\leq \frac{1+\ln n_d}{q_{n_d}\cdot p_{\leq \epsilon}}\\
        &\leq (1+o(1))\cdot e \cdot n_d \cdot \ln n_d \cdot \frac{C}{C-M}.
    \end{align*}

    Since each lower-level call uses $c_{\text{cap}}\Brackets{n_c,R_{\text{GPOM}}}$ fitness evaluations, we obtain the claimed upper bound on the expected first hitting time.
\end{proof}

\begin{proof}
    In order to prove the theorem, we use the multiplicative drift theorem to bound the number of upper-level iterations until the joint objective is at most~$\epsilon$.

    For a fixed discrete offspring $x$, the lower-level objective is a translated \Sphere function plus the constant $w \sum_{i=1}^{n_d}\Brackets{1-x_i}$.
    This constant does not affect selection or step-size adaptation.
    Under the initialization \textsc{InitGen}, each call starts from $y_0$ with step size $\sigma_{\text{init}}=1$, and $\NormEucl{y_0-y^*(x)}\leq R_{\text{GPOM}}$.
    Applying \Cref{lem:akimoto-lower-bound-prob} with $d=n_c$ therefore shows that the returned continuous error is at most $\epsilon$ with probability at least $p_{\le\epsilon}=1-\frac{M}{C}>0$ uniformly over all previous histories and discrete offspring.

    For each integer $u \geq 0$, let $\Brackets{x_u,y_u}$ denote the incumbent after $u$ completed upper-level iterations.
    Write $F_u:= f\Brackets{x_u,y_u}$ and define
    \begin{align*}
        U := \inf\CurlyBrackets{u \geq 0 \mid F_u \leq \epsilon}.
    \end{align*}
    Thus, $U$ counts the completed upper-level iterations until the incumbent first has objective value at most $\epsilon$.
    The target may already be evaluated during the final lower-level call.
    Since each call uses $c_{\text{cap}}\Brackets{n_c,R_{\text{GPOM}}}$ fitness evaluations, we have
    \begin{align}
        \Ex{T_\epsilon} \leq \Ex{U} \cdot c_{\text{cap}}\Brackets{n_c,R_{\text{GPOM}}}.\label{eq:sep:first-hitting-time}
    \end{align}

    To bound the number of upper-level iterations $U$, we apply \Cref{thm:prelim:mult} to the potential process
    \begin{align*}
        G_u := g\Brackets{x_u,y_u} = \minWBrackets{n_d, \left \lceil \frac{F_u-\epsilon}{w}\right\rceil}.
    \end{align*}
    Then $G_u \in \CurlyBrackets{0,\dots, n_d}$, and the elitist selection of the algorithm ensures that $G_{u+1} \leq G_u$.
    Moreover, $G_u = 0$ if and only if $F_u \leq \epsilon$.
    Thus, $U$ is precisely the first hitting time of zero by the potential process.

    Fix an iteration $u$, and let $\mathcal{F}_u$ denote the history of the algorithm before generating the discrete offspring in that iteration.
    Condition on a history with $G_u > 0$.
    The current solution and the quantities defined below are then fixed.

    Let $I := \CurlyBrackets{i \in \firstInts{n_d} \mid x_{u,i}=0}$ and $k := \lvert I\rvert$.
    Thus, the discrete contribution to $F_u$ is $wk$.
    Since the continuous contribution is nonnegative and $\epsilon<w$, we have
    \[
    \frac{F_u-\epsilon}{w}
    \ge k-\frac{\epsilon}{w}>k-1.
    \]
    Taking the ceiling and using $k\le n_d$ yields $G_u\ge k$.
    The probability that a specific bit flips and all other bits remain unchanged is given by $q_{n_d} := \frac{1}{n_d}\Brackets{1-1/n_d}^{n_d-1}$.

    We show that, whenever $u < U$, the conditional expected decrease satisfies
    \begin{align}
        \Ex{G_u-G_{u+1} \mid \mathcal{F}_u} \geq q_{n_d} p_{\leq \epsilon} G_u. \label{eq:thm-up-bilevel-drift}
    \end{align}
    We prove this by dividing the analysis into two cases according to whether the discrete part is optimal.

    \textbf{Case I}: $k\geq 1$.
    The first case applies as long as the discrete part is not fully optimized.

    For each $i \in I$, let $A_i$ denote the event that only the bit $i$ flips and the continuous optimization on the lower level results in an error of at most $\epsilon$.
    The probability that only the bit at index $i$ flips is given by $q_{n_d}$ and \Cref{lem:akimoto-lower-bound-prob} lower-bounds the probability of an error of at most $\epsilon$ by $p_{\leq \epsilon}$ for every possible discrete offspring~$x'_u$.
    Thus, the probability of event $A_i$ can be lower-bounded by $\Prob{A_i\mid \mathcal{F}_u}\geq q_{n_d}p_{\leq \epsilon}$.
    Conditional on the event~$A_i$, and using that $\epsilon < w$, we obtain
    \begin{align*}
        f\Brackets{x'_{u},y'_u} \leq w (k-1) + \epsilon < wk  \leq f\Brackets{x_u,y_u}.
    \end{align*}

    Thus, conditional on $A_i$, the offspring is accepted.
    Since its continuous error lies in $[0,\epsilon]$, we have
    \[
    w(k-1)\le F_{u+1}\le w(k-1)+\epsilon.
    \]
    Consequently,
    \[
    k-2
    <
    k-1-\frac{\epsilon}{w}
    \le
    \frac{F_{u+1}-\epsilon}{w}
    \le k-1.
    \]
    Taking the ceiling and using $k-1<n_d$ yields
    $G_{u+1}=k-1$.
    Hence, the potential decreases by $G_u-k+1$.

    For all $i \in I$, the events $A_i$ are pairwise disjoint.
    Moreover, the elitist selection of the algorithm ensures that the potential cannot increase on any other event.
    Summing the contributions of the events $A_i$ and using $k\geq 1$ and $G_u\geq k$ in the second inequality, we obtain
    \begin{align*}
        \Ex{G_u-G_{u+1} \mid \mathcal{F}_u} &\geq q_{n_d}p_{\leq \epsilon} \sum_{i\in I}\Brackets{G_u-k+1}\\
        &= q_{n_d}p_{\leq \epsilon}\Brackets{k\Brackets{G_u-k}+k}\\
        &\geq q_{n_d}p_{\leq \epsilon}G_u.
    \end{align*}
    The last inequality follows from
    \begin{align*}
        k\Brackets{G_u-k}+k - G_u = (k-1)\Brackets{G_u-k} \geq 0.
    \end{align*}
    This establishes \Cref{eq:thm-up-bilevel-drift} and concludes Case~I.

    \textbf{Case II}: $k= 0$.
    The discrete part is already optimal, so the current fitness consists entirely of the continuous error.
    Since $G_u > 0$, this error is larger than $\epsilon$.
    Let $A$ denote the event that no bit flips and the continuous optimization on the lower level returns a solution with a continuous error of at most $\epsilon$.
    Thus, conditional on $A$, the offspring has fitness at most $\epsilon$, is accepted, and reduces the potential from~$G_u$ to zero.

    Using that no bit flips with probability $p_0 = \Brackets{1-1/n_d}^{n_d}$, the probability of $A$ is bounded below by $\Prob{A\mid \mathcal{F}_u} \geq p_0p_{\leq \epsilon}$.

    With $p_0 = \Brackets{n_d-1}q_{n_d} \geq q_{n_d}$ and since, conditional on $A$, it holds that $G_u-G_{u+1} = G_u$, we get
    \begin{align*}
        \Ex{G_u-G_{u+1} \mid \mathcal{F}_u} &\geq \Prob{A\mid \mathcal{F}_u} \Ex{G_u-G_{u+1} \mid \mathcal{F}_u,A}\\
        &\geq p_0p_{\leq \epsilon} G_u\\
        &\geq q_{n_d}p_{\leq \epsilon}G_u,
    \end{align*}
    which establishes the drift condition in \Cref{eq:thm-up-bilevel-drift} and concludes Case~II.

    The two cases establish~\Cref{eq:thm-up-bilevel-drift} in every history with positive potential.
    Taking conditional expectations given $G_0,\ldots,G_u$ yields the drift condition required by the multiplicative drift theorem.

    Applying the multiplicative drift theorem with $g\Brackets{x_0,y_0} \leq n_d$, $q_{n_d}\geq1/\Brackets{en_d}$, and $p_{\leq \epsilon}\geq 1-M/C$ results in
    \begin{align*}
        \Ex{U} &\leq \frac{1+\ln n_d}{q_{n_d}\cdot p_{\leq \epsilon}}\\
        &\leq \frac{e \cdot n_d\cdot \Brackets{1+\ln \Brackets{n_d}}}{(1-M/C)} \\
        &\leq (1+o(1))\cdot e \cdot n_d \cdot \ln n_d \cdot \frac{C}{C-M}.
    \end{align*}
    Here, $o(1)$ refers to $n_d \rightarrow \infty$.

    Inserting this into \Cref{eq:sep:first-hitting-time} concludes the proof of the theorem.\qedhere
\end{proof}

We now specialize the bounds to \SphereOneMax, Type-II,
and \WPMCProximity, and extend the argument to a special
case of Type-I and Type-III interactions that include masking effects.

In the following corollaries,
use \textsc{InitGen} and \textsc{StopCritCap} with budget
$c_{\text{cap}}\Brackets{n_c,R}$, where $R$ is specified separately
for each problem.
Assume $\epsilon\in\Brackets{0,\minWBrackets{1,w}}$, let $M$ be the constant from \Cref{lem:akimoto-lower-bound-prob} and fix $C\ge2$ with $C>M$,
The first hitting time $T_\epsilon$, dimensions, and convention for
initialization are as in \Cref{thm:bilevel:gen:lb,thm:bilevel:gen:ub}.
All asymptotic statements refer to $n_d\to\infty$.

\begin{corollary}
\label{cor:bilevel:sphereonemax}
\label{cor:bilevel:type2}
\label{cor:bilevel:wpmcprox}
    Consider \SphereOneMax, a Type-II interaction problem, or \WPMCProximity with the parameters in \Cref{tab:functions:gen-sphere-lin}.
    Define $R:=R_{\text{GPOM}}$ as in~\Cref{eq:bilevel:benchmark-proofs:def-R} with the corresponding mapping $y^*$.
    Under the assumptions stated above, the expected first hitting time of the \oneoneEAESBL is bounded from below by
    \begin{align*}
        \Ex{T_\epsilon} &\geq (1-o(1))\cdot e \cdot n_d \cdot \ln\Brackets{n_d} \cdot c_{\text{cap}}\Brackets{n_c,R},
    \end{align*}
    and above by
    \begin{align*}
        \Ex{T_\epsilon} &\leq (1+o(1))\cdot e \cdot n_d \cdot \ln\Brackets{n_d} \cdot \frac{C}{C-M} \cdot c_{\text{cap}}\Brackets{n_c,R}.
    \end{align*}
\end{corollary}

\begin{proof}
    By \Cref{tab:functions:gen-sphere-lin}, \SphereOneMax, the interaction problems of Type-II and \WPMCProximity belong with the stated parameters to the function class \GenSphereLinear.
    Thus, by applying \Cref{thm:bilevel:gen:lb,thm:bilevel:gen:ub}, the corollary follows.
\end{proof}

The Type-I and Type-III objectives do not belong to the \GenSphereLinear function class, because the discrete variables determine which continuous coordinates are active.
Nevertheless, we obtain the same run time bounds for the following restricted benchmarks by projecting each lower-level run onto its active coordinates.
For Type-III, the restriction retains both the masking of continuous coordinates and the dependence of their conditional optima on other bits, while ensuring that every inactive coordinate contributes exactly $w$ to the objective.

\begin{corollary}
    \label{cor:bilevel:type1-type3}
    \label{prop:bilevel:type1}
    \label{cor:bilevel:type3}
    Fix the constant $w>0$ and let $n_d=n_c=N\geq2$.
    Consider the Type-I function with $b_i^*=\sqrt{w}$ for every $i\in\firstInts{N}$, or the Type-III function with
    \[
        \phi_i^*(x)=\sqrt{w}\,\Brackets{1-2x_{\pi(i)}},
    \]
    where $\pi$ is a fixed cyclic permutation of~$\firstInts{N}$.
    Choose $R$ satisfying~\Cref{eq:bilevel:benchmark-proofs:def-R}, using $y^*(x)=\sqrt{w}\,\mathbf{1}_{n_c}$ for Type-I and
    $y^*(x)=\phi^*(x)$ for Type-III.
    Under the initialization, budget, and parameter assumptions of \Cref{thm:bilevel:gen:lb,thm:bilevel:gen:ub}, with this choice of $w$ and $R$, the expected first hitting time of the \oneoneEAESBL satisfies
    \begin{align*}
        \Ex{T_\epsilon} &\geq (1-o(1))\cdot e \cdot n_d \cdot \ln\Brackets{n_d} \cdot c_{\text{cap}}\Brackets{n_c,R},
    \end{align*}
    and
    \begin{align*}
        \Ex{T_\epsilon} &\leq (1+o(1))\cdot e \cdot n_d \cdot \ln\Brackets{n_d} \cdot \frac{C}{C-M} \cdot c_{\text{cap}}\Brackets{n_c,R}.
    \end{align*}
\end{corollary}

\begin{proof}[Proof sketch]
    Both objectives decompose into $w\sum_{i=1}^N\Brackets{1-x_i}$ and a translated \Sphere function on the active coordinates.
    Since the projected initial distance is at most $R$, \Cref{lem:akimoto-lower-bound-prob} provides the required success guarantee uniformly over all active dimensions.
    The discrete comparison and drift arguments from the proofs of \Cref{thm:bilevel:gen:lb,thm:bilevel:gen:ub} therefore apply unchanged.
\end{proof}

\begin{proof}
    \textbf{Type-I.}
    For $x \in \CurlyBrackets{0,1}^{N}$, let $A(x) := \CurlyBrackets{i \in \firstInts{N} \mid x_i = 1}$ denote the active coordinates.
    Then, the joint objective decomposes as
    \begin{align*}
        f_I(x,y) = w \sum_{i= 1}^N\Brackets{1-x_i} + \sum_{i \in A(x)}\Brackets{y_i - \sqrt{w}}^2.
    \end{align*}
    In particular, $f_I(x,y) \leq \epsilon < w$ implies $x = 1^N$.
    The discrete generation process and the fixed lower-level budget therefore satisfy the conditions used in the proof of \Cref{thm:bilevel:gen:lb}, which establishes the lower bound.

    For the upper bound, fix a discrete offspring $x$.
    Restricted to its active coordinates and translated by $\sqrt{w}$, the lower-level process follows the \oneoneES on a \Sphere function of dimension $d:= \lvert A(x)\rvert$.
    Indeed, the inactive coordinates do not affect either selection or step-size adaptation.
    Under the lower-level initialization \textsc{InitGen}, the projected initial point~$z_0$ satisfies
    \begin{align*}
        \NormTwo{z_0}=\sum_{i\in A(x)}\Brackets{y_{0,i}-\sqrt{w}}^2\le\NormTwo{y_0-\sqrt{w}\,\mathbf{1}_{n_c}}\le R^2.
    \end{align*}
    \Cref{lem:akimoto-lower-bound-prob} applies uniformly to every $d\in\{0,\ldots,N\}$.
    For $d=0$, the continuous error is identically zero.
    Thus, conditional on any previous history and any discrete
    offspring, the lower-level run returns a point with continuous
    error at most $\epsilon$ with probability at least $p_{\le\epsilon}=1-M/C$.

    The objective has the same discrete contribution and a
    nonnegative continuous error satisfying the same conditional
    success bound as in \Cref{thm:bilevel:gen:ub}.
    Therefore, the potential and the two-case drift argument from
    its proof apply unchanged and establish the upper bound.

    \textbf{Type-III.}
    Let $A(x):=\{i:x_i=1\}$.
    Since $\phi_i^*(x)^2=w$ for every $i\in \firstInts{N}$ and $x \in \CurlyBrackets{0,1}^N$,
    the objective decomposes as
    \[
        f_{III}(x,y)
        =w\sum_{i=1}^N\Brackets{1-x_i}
        +\sum_{i\in A(x)}\Brackets{y_i-\phi_i^*(x)}^2.
    \]
    For each fixed $x$, the second term is a translated \Sphere
    function on the active coordinates, with initial distance
    at most $R$.
    Hence, the discrete comparison and active-coordinate
    projection arguments of
    the Type-I case apply unchanged.
\end{proof}

\section{Conclusion}
\label{sect:conclusion}
We analyzed how interactions between discrete and continuous variables affect the performance of integrated and bilevel evolutionary algorithms on mixed discrete-continuous benchmarks.
On \WPMCProximity, which contains pairwise interactions, our experiments showed that the \oneoneEAESHB frequently enters states near suboptimal conditional continuous optima with small step sizes.
For initial states satisfying the analyzed trap conditions, we proved an exponential lower bound on the expected time to reach the target accuracy.
To address this difficulty, we introduced the \oneoneEAESBL, which optimizes the continuous variables for a proposed discrete configuration before deciding whether to accept the joint offspring.
For the configuration with a fixed continuous starting point and a suitable fixed budget per lower-level call, we obtained matching asymptotic bounds on the expected number of fitness evaluations needed to reach objective value at most $\epsilon$ on the analyzed benchmarks, including \WPMCProximity: $\ThetaWBrackets{n_d \cdot \log n_d \cdot n_c \cdot \logWBrackets{R/\sqrt{\epsilon}} }$, where $n_d$ is the number of discrete variables, $n_c$ is the number of continuous variables, and~$R$ bounds the distances from the fixed continuous starting point to all relevant conditional optima.

It remains to determine how an initial continuous warm-starting phase, initialization of each lower-level call at the current
continuous solution, or adaptive stopping criteria affect the expected run time and its dependence on~$R$.
This includes whether any of those configuration options results in a non-quadratic expected run time of the bilevel algorithm.
Extending the analysis beyond the restricted Type-I and Type-III benchmarks is another open question.
A further direction is to extend the analysis to mixed discrete-continuous applications, such as network connection problems in which discrete construction decisions interact with continuous operating parameters.

\paragraph{Use of generative AI}
We used OpenAI's ChatGPT Education to generate experimental code from detailed author-specified algorithms, parameters, experiments, and plotting requirements, and to check mathematical statements for potential errors.
It was not used to develop research ideas or any part of the proofs.
The authors remain fully responsible for all content, code, and results.
\begin{acks}
We used OpenAI's ChatGPT Education to improve the language and rephrase individual sentences written by the authors.
Its use in the research process is disclosed in the paragraph on generative AI above.
The authors remain fully responsible for the wording and content of the manuscript.
\end{acks}

\newpage

\begingroup
\interlinepenalty=10000

\endgroup

\end{document}